\documentclass[conference]{IEEEtran}
\IEEEoverridecommandlockouts
\usepackage{cite}
\usepackage{amsmath,amssymb,amsfonts}
\usepackage{algorithmic}
\usepackage{graphicx}
\usepackage{textcomp}
\usepackage{xcolor}
\usepackage{hyperref}
\def\BibTeX{{\rm B\kern-.05em{\sc i\kern-.025em b}\kern-.08em
    T\kern-.1667em\lower.7ex\hbox{E}\kern-.125emX}}
\begin{document}

\title{{Impact, Causation and Prediction of Socio-Academic and Economic Factors in Exam-centric Student Evaluation Measures using Machine Learning and Causal Analysis }  \\
}

\author{
\IEEEauthorblockN{Md. Biplob Hosen\textsuperscript{1,2}, Sabbir Ahmed\textsuperscript{3}, Bushra Akter\textsuperscript{1}, Mehrin Anannya\textsuperscript{1}}

\IEEEauthorblockA{\textsuperscript{1}\textit{Institute of Information Technology}, \textit{Jahangirnagar University}, Savar, Dhaka, Bangladesh }
\IEEEauthorblockA{\textsuperscript{2}\textit{Department of Information Systems}, \textit{University of Maryland Baltimore County}, Maryland, United States}
\IEEEauthorblockA{\textsuperscript{3}\textit{Erik Jonsson School of Engineering and Computer Science, University of Texas}, Dallas, United States}

biplob.hosen@juniv.edu, sabbir.ahmed2@utdallase.edu, bushrajuit@gmail.com, mehrin.anannya@juniv.edu 
}

\maketitle
\begin{abstract}
Understanding socio-academic and economic factors influencing students' performance is crucial for effective educational interventions. This study employs several machine learning techniques and causal analysis to predict and elucidate the impacts of these factors on academic performance. We constructed a hypothetical causal graph and collected data from 1,050 student profiles. Following meticulous data cleaning and visualization, we analyze linear relationships through correlation and variable plots, and perform causal analysis on the hypothetical graph. Regression and classification models are applied for prediction, and unsupervised causality analysis using PC, GES, ICA-LiNGAM, and GRASP algorithms is conducted. Our regression analysis shows that Ridge Regression achieve a Mean Absolute Error (MAE) of 0.12 and a Mean Squared Error (MSE) of 0.024, indicating robustness, while classification models like Random Forest achieve nearly perfect F1-scores. The causal analysis shows significant direct and indirect effects of factors such as class attendance, study hours, and group study on CGPA. These insights are validated through unsupervised causality analysis. By integrating the best regression model into a web application, we are developing a practical tool for students and educators to enhance academic outcomes based on empirical evidence.

\end{abstract}


\section{Introduction}
Integrating technology into education systems has developed new frameworks to address the persistent challenge of student attrition, a concern for both educational institutions and nations. Students often disengage and ultimately leave their studies due to various academic difficulties, influenced by factors such as financial constraints, lack of motivation, insufficient study time, or general disinterest in the curriculum \cite{l1}. Conventional educational models, often deterministic and lacking advanced analytical tools, inadequately address the complexities of student performance, offering little beyond generalized approaches that fail to meet the unique needs of each student and are ineffective at predicting educational outcomes. In contrast, machine learning (ML)-based approaches have significantly transformed the understanding and prediction of student performance by leveraging diverse data sources, including numerical data from surveys, video feeds, and behavioral analyses. These methodologies extend across various domains of ML and deep learning (DL), utilizing image, numeric, video, and even smartwatch data to analyze and predict educational outcomes \cite{l2}.

Studies focused on predicting students' semester grade point averages (SGPA) using ML techniques reveal limitations, such as a narrow focus on specific courses and challenges in incorporating diverse assessment factors, particularly during the COVID-19 pandemic \cite{l3, l4}. Additionally, models that identify students at risk of academic failure often rely on limited input features and may not fully capture all variables influencing performance \cite{l5, l8}. Research on cumulative GPA (CGPA) and final semester grades emphasizes demographic and academic data but is constrained by issues of data quality and institutional variability \cite{l6, l7, l9, l10}.

A critical evaluation of these approaches reveals a common limitation: an excessive focus on algorithmic improvements without sufficient consideration of real-life factors, such as social, academic, and economic situations. Therefore, key factors from the literature that mostly affect academic performance have been identified. Table ~\ref{table:factor} effectively illustrates these findings.

\begin{table*}[htbp]

  \centering
  \caption{ Factors Influencing Student Performance and Their Acronyms with References}
    \begin{tabular}{lllll p{4.03cm}}
    \hline
   Factors & Acronym & Relation [Ref] & Factors & Acronym & Relation [Ref]
 \\
    \hline
    Department/Institute &Dept &Less effect & S.S.C Result(GPA) &SSC & Motivation\cite{lt4}
\\

Year/Semester & Sem & No effect & H.S.C Result (GPA) &HSC & Motivation and confidence\cite{lt4}
\\

Gender & G & Sometimes & Father Education 	 & F.edu & Influential\cite{lt5}
\cite{lt6}
\\
Father Job & F.job & Influential\cite{lt5}
\cite{lt6}
 &Mother Education &M.edu & Influential\cite{lt5}
\cite{lt6}
\\

Mother Job & M.job & Influential\cite{lt5}
\cite{lt6}
&Major Illness & Ill & Impairing\cite{lt7}
\\
 Attendance in Class &Att &Consistency\cite{lt9}&Group Study	&GC & Collaborative\cite{lt18}\cite{lt19}\\
Study Hour (in a week) &SH&Discipline &Sport/Cultural &Sp &Enriching \cite{lt20,lt21,lt22}\\

Political Involvement &PIn &Distraction\cite{lt17}&
Internet Facilities &Int &Distracting and resources
\cite{lt8,lt14,lt15,lt16}
\\

Getting Any Scholarship &Sch &Confidence\cite{lt10}\cite{lt11} &Hostel Staying	 &HS &Independence\cite{lt23}\cite{lt24}\cite{lt25}\\

Self-Income (in BDT) &SI &Confidence\cite{lt27} &Relational Status &Rel &Distracting and supportive\cite{lt12}\cite{lt13}\\
Communication Skill &Com &Influence\cite{lt26} &Confidence &Con &Correlation\cite{lt27}\\

    \hline
    \end{tabular}
  \label{table:factor}%
\end{table*}%

Moreover, the distinction between correlation and causality is crucial in this context. Causal inference provides a framework for understanding the directional impacts of various factors on student performance, which is essential for developing effective interventions to improve academic outcomes. However, aligning initial causal hypotheses with empirical data remains challenging. Discrepancies between hypothesized causality and observed data highlight the complexity of the educational landscape and suggest the presence of significant, yet unidentified, influencing factors.

Recognizing these challenges, this research emphasizes the paramount importance of causality analysis in educational data science. Causality, unlike mere correlation, offers insights into the effects of various factors on student performance, enabling targeted interventions and strategies. This work presents a comprehensive framework that includes the development of an Initial Hypothetical Causal Graph (GCM), meticulous data collection, and sophisticated data cleaning and visualization techniques. By exploring the relationships between CGPA and various factors through correlation and multifactorial plots, the research aims to identify the causes of changes in CGPA.

The culmination of this research is the integration of the most effective predictive model into a web application, providing an accessible and practical tool for educational stakeholders. This addresses the critical need for a comprehensive survey with sufficient data to establish relationships experimentally and verify their alignment with existing literature. By offering an open-source system for public use, the goal is to bridge the gap between advanced ML methodologies and essential socio-academic and economic factors, fostering a more informed, effective, and personalized approach to educational evaluation.


\section{Materials and Methodology}

\begin{figure}[htbp]
	\centering
		\includegraphics[width=\linewidth]{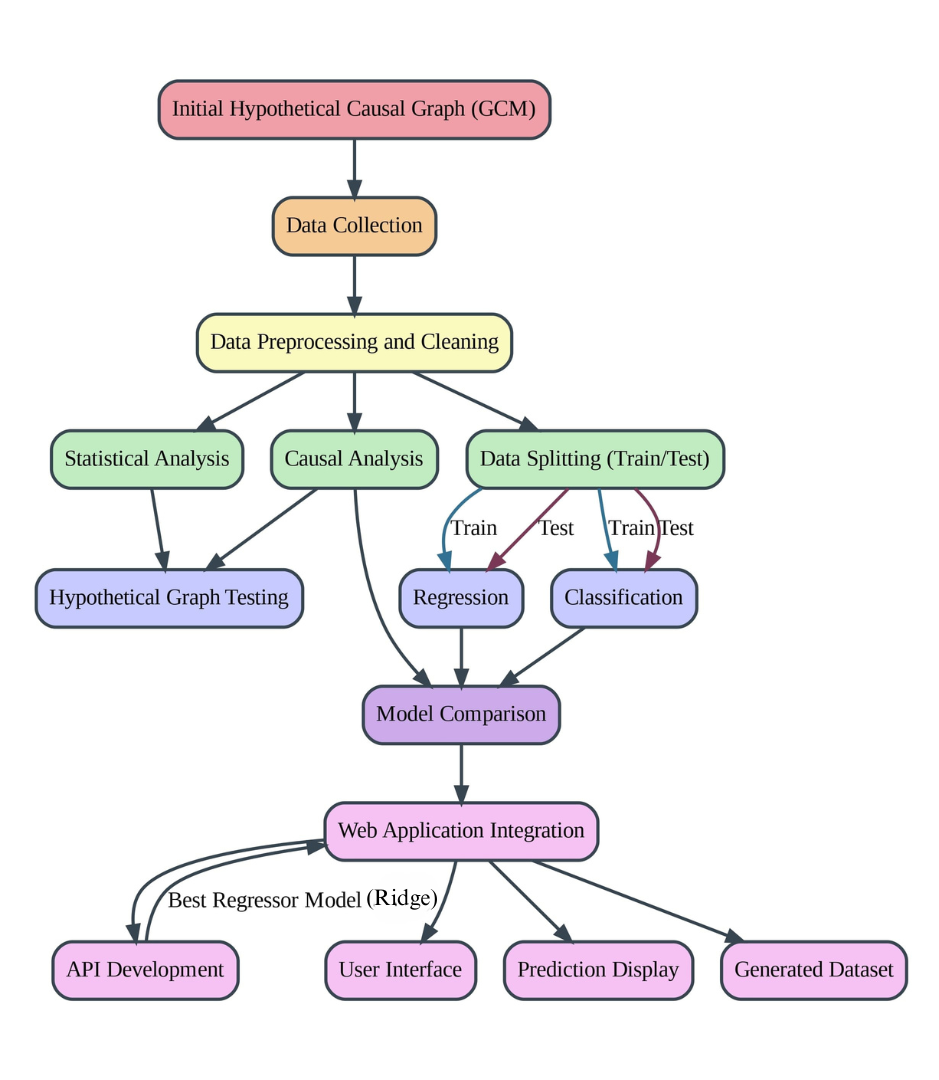}
	\caption{Overall workflow diagram of the study. }
	\label{workflow}
\end{figure}

The work procedure described in Figure~\ref{workflow} begins with the formulation of an initial hypothesis graph. This is followed by the stages of data collection, preprocessing, and cleaning. The preprocessing phase is subdivided into statistical analysis, causal analysis, and data splitting for testing and training. Hypothesis graph clustering is used during the statistical and causal analysis stages, while regression and classification techniques are applied after the data split. The strategy concludes with model comparison and the integration of a web application, ensuring a comprehensive and effective system implementation.

\subsection{Data Collection}
The data collection phase involves a dataset of 1,050 student profiles, capturing a wide range of academic and personal attributes essential for understanding the factors influencing CGPA. This dataset provides a comprehensive view of student backgrounds, behaviors, and circumstances, enabling a thorough analysis of academic performance. Our analysis is guided by previous literature, as summarized in Table~\ref{table:factor}.

\subsection{Data Cleaning, Processing and Statistical Analysis}

We begin our analysis with a thorough exploration of the dataset, using various descriptive statistics and visualizations such as scatter plots, histograms, and box plots. For instance, scatter plots highlights positive correlations between class attendance and CGPA, as well as study hours and CGPA, underscoring the significance of these factors in academic achievement. Additionally, our statistical analysis involve examining the relationships between different factors to construct an initial hypothesis graph. This graph is weighted according to the strength of the correlations, providing a preliminary model for causality analysis.

 \subsection{Prediction by Regression}

The primary goal of this study is to predict students' CGPA using advanced regression techniques. Linear Regression serves as a baseline, while Robust, Ridge, Lasso, and Elastic Net regressions help manage outliers, overfitting, and feature selection. More complex models, such as Stochastic Gradient Descent, Random Forest, SVM, and Artificial Neural Networks (ANNs), are used to capture non-linear and complex relationships in the data, improving prediction accuracy. ANNs, in particular, captures complex, non-linear relationships with multiple layers and neurons, with various setups tested for optimal performance.

\subsection{Prediction by Classification}

For predicting students' grade, a variety of classification algorithms are employed. Logistic Regression handles binary classification, while Stochastic Gradient Descent and Ridge Classifier improve efficiency and performance, respectively. Ensemble methods like Random Forest and AdaBoost enhanced accuracy, and XGBoost provide powerful iterative refinement. K-Nearest Neighbors (KNN) and Decision Trees offer simplicity in classification, while the Multi-layer Perceptron (MLP) Classifier uses neural networks to capture complex patterns in the data.

\subsection{Unsupervised Causality Analysis}

This study uses causal analysis algorithms like PC, GES, ICA-LiNGAM, and GRASP to identify cause-effect relationships in observational data. The PC algorithm constructs causal graphs by testing conditional independencies, while GES uses a score-based approach to search for the best-fitting graph. ICA-LiNGAM focuses on linear non-Gaussian acyclic models, and GRASP combines structural learning with causal discovery to refine the results. These methods collectively reveal dependencies and interactions among features.

 \subsection{Web Application Integration with the Best Model}

Finally, the methodology includes web application design. The system design follows a structured approach across different layers, as detailed in Table \ref{table:tech_layers}. The front end is developed using React.js, HTML, and CSS for a responsive user interface, while the back end utilizes Node.js, Express.js, and JWT for secure server-side application development. MySQL and Sequelize facilitate efficient database operations, managed through XAMPP for local development. The system's data flow is organized using a microservices architecture to ensure effective data processing. Emphasizing continuous improvement, the development process incorporates agile methodologies and automated testing to maintain code quality. Deployment to a cloud-based environment, coupled with DevOps practices, ensures scalability and efficient monitoring of system performance, resulting in a robust web application tailored for students and educators.
\begin{table}[htbp]
  \centering
  \caption{Technologies used in each layer of the architecture}
  \begin{tabular}{|c|c|}
    \hline
    \textbf{Layer} & \textbf{Technologies} \\
    \hline
    \textbf{Front End} & ReactJS, HTML, CSS \\
    \hline
    \textbf{Back End} & Node.js, Express.js, JWT \\
    \hline
    \textbf{Database} & MySQL, Sequelize, XAMPP \\
    \hline
  \end{tabular}
  \label{table:tech_layers}
\end{table}

\section{Experimental Results}
\subsection{Statistical Analysis}

\begin{figure*}[htbp]
	\centering
		\includegraphics[width=0.45\linewidth]{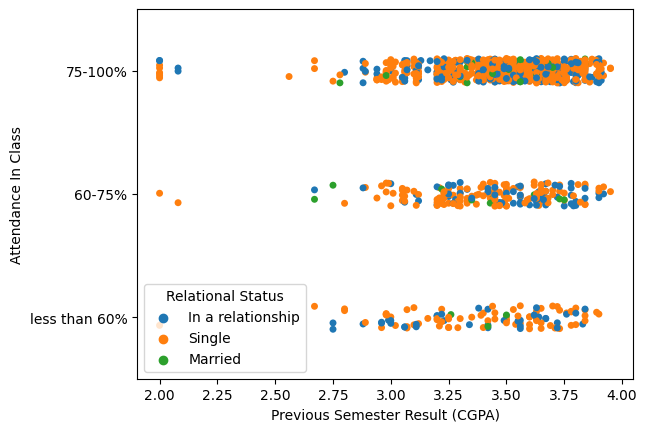} 
  (a)
  \includegraphics[width=0.45\linewidth]{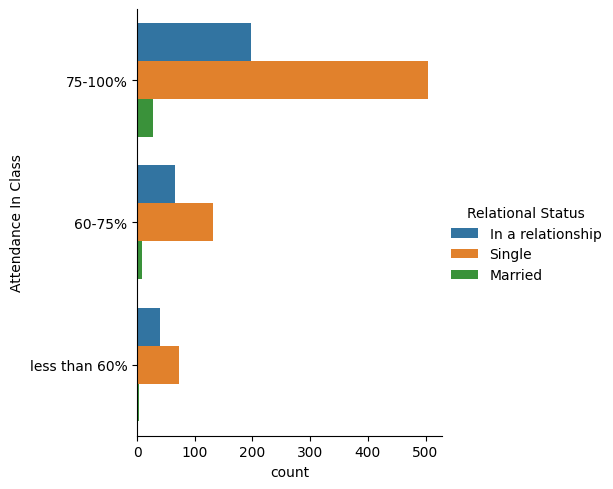}
  (b)
 
	\caption{statistical analysis of various filter: (a) Correlation between attendance, CGPA, and relationship status. (b) Attendance count for different relationship status.}
    \label{SA}
\end{figure*}

The statistical analysis detailed in Figure~\ref{SA} examines various factors influencing CGPA. Part (a) shows a significant correlation between attendance, relationship status, and CGPA, with students maintaining 75\% to 100\% attendance clustering around CGPAs of 3.75 to 4.0, while those with less than 60\% attendance exhibits a wider range of lower CGPAs. Additionally, part (b) illustrates that single students are more consistent in class attendance compared to those in relationships or married, and class attendance positively impacts academic performance. We also analyze that confidence in academic abilities correlates with higher CGPAs, while confidence in non-academic areas can lead to lower academic performance. Furthermore, studying 10–14 hours per week produces the best CGPA outcomes, with fewer study hours often leading to lower CGPAs, and excessive study hours (over 14) negatively impacting both health and performance. A balanced approach to attendance, confidence, and study hours is essential for academic success.

\subsection{Regression Results}

The results of our study highlight the effectiveness of various regression and classification models in predicting students' CGPA. In the regression analysis, performance metrics are detailed in Table~\ref{regression}. Linear Regression serves as a baseline, showing modest performance with a Mean Absolute Error (MAE) of 0.120 and a Mean Squared Error (MSE) of 0.024. The R-squared value is slightly negative, indicating that the model does not fit the data well. Robust Regression performs poorly, with a high MAE of 0.615 and an MSE of 10.762, suggesting it is significantly affected by outliers. Ridge Regression slightly improves upon Linear Regression, with an MAE of 0.120 and an R-squared value close to zero, indicating minimal improvement. Lasso and Elastic Net Regression produce similar results to Ridge Regression, with only minor differences in performance metrics.

\begin{table*}[htbp]

  \centering
  \caption{ Regression model performance matrices}
    \begin{tabular}{llllll}
    \hline
   \textbf{Model} & \textbf{MAE} & \textbf{MSE} & \textbf{RMSE} & \textbf{R\textsuperscript{2} Score} & \textbf{Cross Validation} 

 (\%) \\
    \hline
    
Linear Regression & 1.204211e-01 & 2.400476e-02 & 1.549347e-01 & -9.559751e-03 & -0.060102 \\
Robust Regression & 6.148482e-01 & 1.076215e+01 & 3.280572e+00 & -4.516200e+02 & -55.587331 \\
Ridge Regression & 1.200878e-01 & 2.370764e-02 & 1.539729e-01 & 2.935949e-03 & -0.059790 \\
Lasso Regression & 1.202154e-01 & 2.373162e-02 & 1.540507e-01 & 1.927581e-03 & -0.034433 \\
Elastic Net Regression & 1.201991e-01 & 2.373139e-02 & 1.540500e-01 & 1.937092e-03 & -0.034433 \\
Stochastic Gradient Descent & 1.814458e+10 & 1.443672e+22 & 1.201529e+11 & -6.071600e+23 & 0.000000 \\
Random Forest Regressor & 1.283269e-01 & 2.913691e-02 & 1.706954e-01 & -2.254007e-01 & 0.000000 \\
SVM Regressor & 1.311040e-01 & 2.915696e-02 & 1.707541e-01 & -2.262442e-01 & 0.000000 \\
Artificial Neural Network & 2.200199e-01 & 6.644527e-02 & 2.577698e-01 & -1.794465e+00 & 0.000000 \\

    \hline
    \end{tabular}
  \label{regression}%
\end{table*}%

\subsection{Classification Results}

For classification tasks aimed at predicting students' grades, various algorithms are evaluated based on their training and testing accuracy and F1-score. The Logistic Regression (LR) classifier achieves a training accuracy of 52.6\% and a test accuracy of 48.8\%, indicating moderate performance with room for improvement. Stochastic Gradient Descent (SGD) performs better, with a test accuracy of 54.1\%, but the F1-score is lower, indicating issues with precision or recall. The Ridge classifier improves performance slightly, with a test accuracy of 55.7\%, while the Passive Aggressive (PA) classifier shows a significant improvement, achieving a test accuracy of 61.98\%. The Decision Tree (DT) classifier and ensemble methods like Random Forest (RF), AdaBoost (AB), and Extreme Gradient Boosting (XGB) perform exceptionally well, with perfect accuracy and F1-scores on both training and test sets, indicating their robustness and high predictive power. The KNN Classifier also performs very well, with a test accuracy of 99.1\%, demonstrating its effectiveness in capturing patterns in the data.

Support Vector machine (SVM) classifier and Linear SVM (LSVC) show strong performance, with SVM achieving a test accuracy of 87.2\%, indicating a good balance between model complexity and performance. The MLP Classifier, a type of neural network, also performs well, with a test accuracy of 98.4\%, showing its capability to handle complex patterns in the data effectively. These results are shown in Table~\ref{classification}.

\begin{table}[htbp]

  \centering
  \caption{ classification model performance analysis}
    \begin{tabular}{p{1cm} p{1.3cm} p{1.3cm} p{1.3cm} p{1.3cm}}
    \hline
  \textbf{ML} & \textbf{Train } & \textbf{Train } & \textbf{Test } & \textbf{Test} 
  \\
  \textbf{Classifier} & \textbf{ Accuracy} & \textbf{ F1-Score} & \textbf{ Accuracy} & \textbf{ F1-Score} 

  \\
    \hline
    LR & 0.525779 & 0.59 & 0.487984 & 0.54 \\
SGD & 0.581083 & 0.27 & 0.540698 & 0.22 \\
Ridge & 0.615325 & 0.31 & 0.556589 & 0.26 \\
PA & 0.669466 & 0.42 & 0.619767 & 0.33 \\
DT & 0.998966 & 1.00 & 0.998450 & 1.00 \\
K-NN & 0.998320 & 1.00 & 0.991085 & 1.00 \\
RF & 0.990438 & 0.997 & 1.00 & 1.00 \\
AB & 0.366068 & 0.02 & 0.357752 & 0.01 \\
XGB & 0.998966 & 1.00 & 0.998450 & 1.00 \\
SVM & 0.893139 & 0.69 & 0.871705 & 0.68 \\
LSVM & 0.695180 & 0.44 & 0.633333 & 0.36 \\
MLP & 0.986691 & 0.96 & 0.984109 & 0.96 \\

    \hline
    \end{tabular}
  \label{classification}%
\end{table}%

\subsection{Causal Models}

\begin{figure}[htbp]
	\centering
		\includegraphics[width=\linewidth]{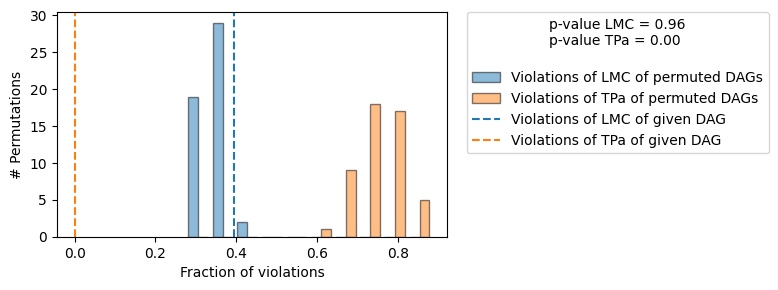} 

	\caption{ Evaluation of the Causal Graph Structure}
    \label{bar}
\end{figure}

\begin{figure*}[htbp]
	\centering
		\includegraphics[width=0.45\linewidth]{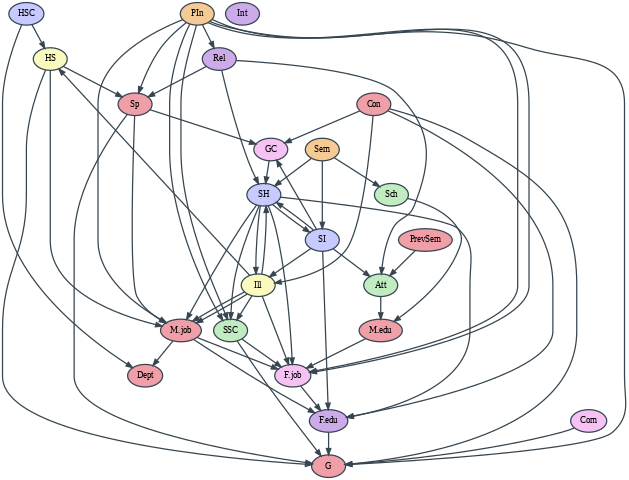} 
  (a)
  \includegraphics[width=0.45\linewidth]{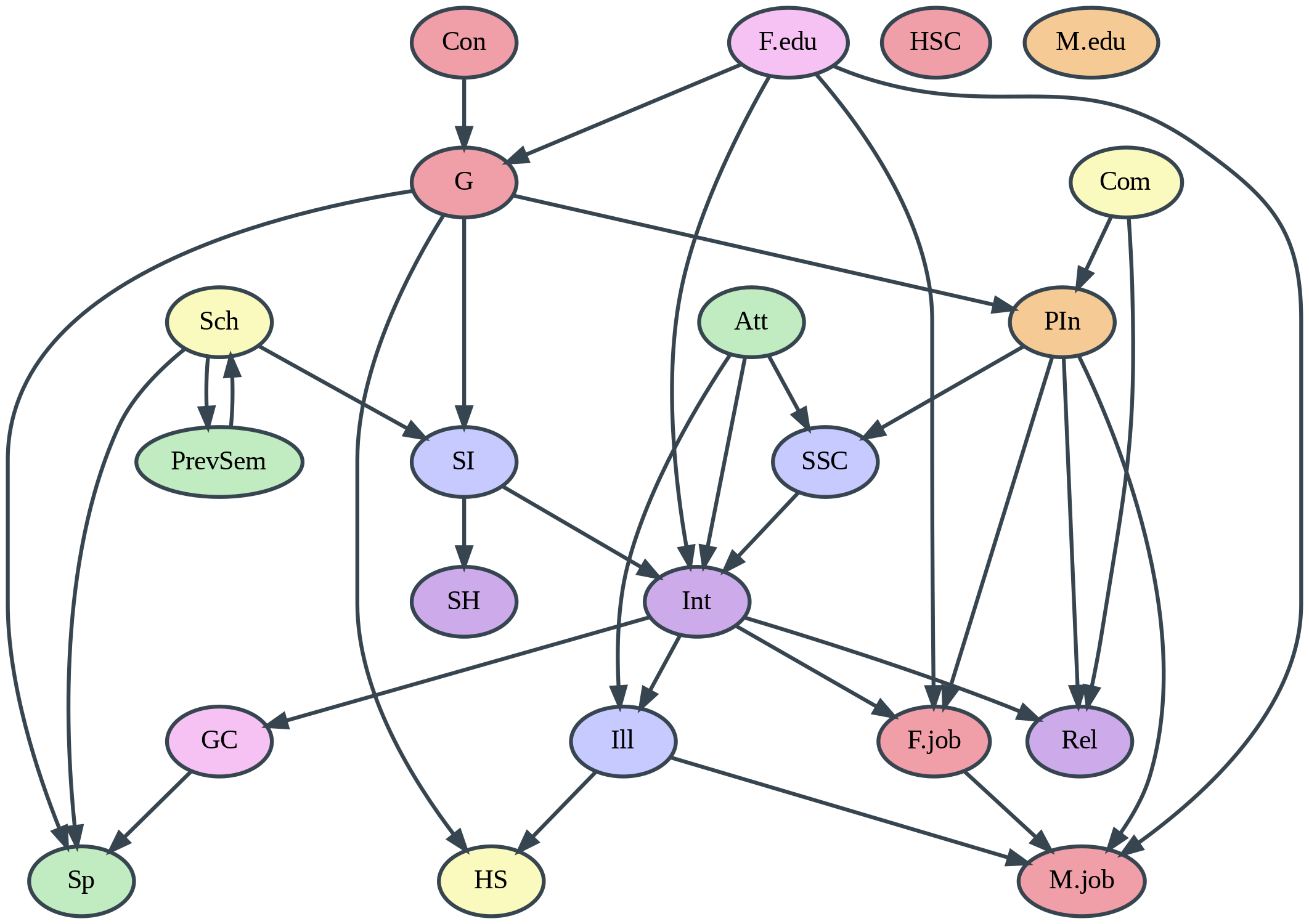}
  (b)
  \includegraphics[width=0.45\linewidth]{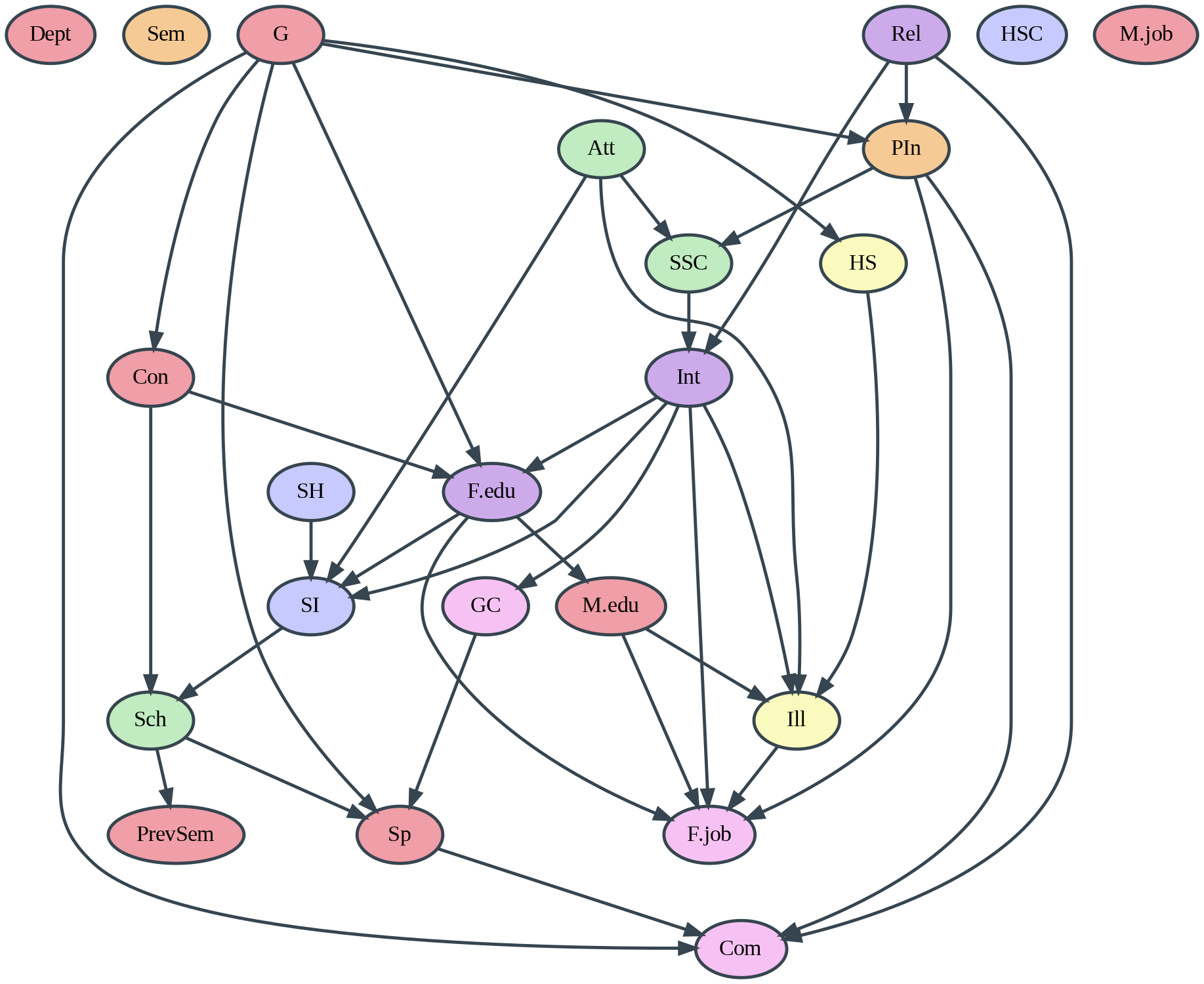}
  (c)
  \includegraphics[width=0.45\linewidth]{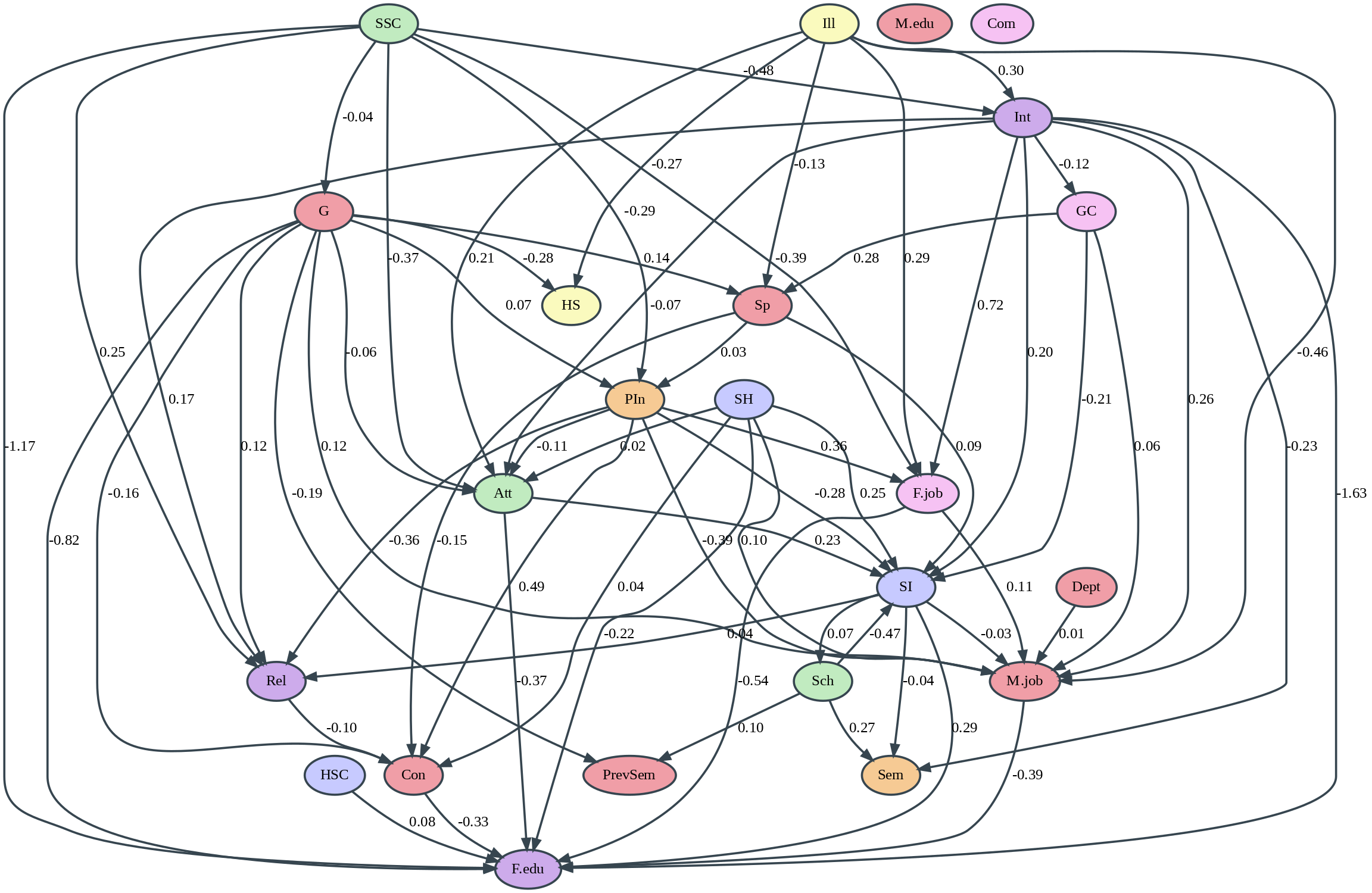}
  (d)
	\caption{various casual model: (a) PC model (b) GES model  (c) GRASP model.
 (d) ICA-LiNGAM model.}
    \label{cau}
\end{figure*}

Causal analysis is conducted using the Generalized Causal Model (GCM) framework, employing algorithms such as PC, GES, GRASP, and ICA-LiNGAM. The results are presented in Figures~\ref{bar} and~\ref{cau}, providing insights into the reliability and accuracy of each model in capturing causal relationships. Figure~\ref{bar} shows a comparison of the permutation and violation factors, indicating the Linear Markov Condition (LMC) holds, but the Total Positivity of Treatment (TPa) assumption is violated.

Figures~\ref{cau}(a)-(d) illustrate the causal graphs generated by the four algorithms. The PC algorithm (Fig.\ref{cau}(a)) produces a relatively complex structure with factors like GC, SH, Att, and SSC, while GES (Fig.\ref{cau}(b)) simplifies the model, focusing on SH, GC, Rel, and Int. GRASP (Fig.\ref{cau}(c)) identifies intricate relationships, highlighting Att, SSC, Int, and Fjob. ICA-LiNGAM (Fig.\ref{cau}(d)) introduces weighted edges, emphasizing the strength of relationships between factors like G, GC, and SH.

Despite differences in complexity, all four algorithms consistently identify GC, Att, and SH as central features. ICA-LiNGAM provides the most detailed causal analysis by revealing the strength of dependencies through weighted edges, offering a deeper understanding of causal relationships within the dataset.

\subsection{Web App Features}

The analysis resulted in the development of a web app interface designed for both students and teachers. Users can securely log in using their email and password. Students can predict their CGPA based on various factors, update their profile information, and receive personalized recommendations for improving their grades. Teachers have additional features, such as live chat for instant communication and the ability to assign scores and provide feedback. The app ensures data accuracy and integrity by allowing users to update their information as needed, promoting effective communication and contributing to academic growth.

\section{Conclusion}

This study highlights the significance of socio-academic and economic factors in predicting and understanding students' academic performance. By leveraging both deterministic models, such as Ridge Regression, and probabilistic models, such as Random Forest, we demonstrate how each approach captures different aspects of the relationship between these factors and CGPA. Ridge Regression offeres stable predictions by penalizing less significant variables, while Random Forest excels in uncovering complex interactions and non-linear relationships. Through causal analysis, we have identified key factors like class attendance, study hours, and group study as having both direct and indirect effects on CGPA. These findings are further validated using unsupervised causality analysis techniques. Our study not only provides valuable insights into the causal dynamics of academic performance but also contributes to the development of a practical tool that integrates machine learning models to aid students and educators in improving academic outcomes. Future research could build on these insights by integrating additional socio-emotional factors, offering a more comprehensive understanding of the drivers behind academic success.

\bibliographystyle{ieeetr}

\bibliography{cas-refs}

\begin{thebibliography}{10}
\bibitem{lt25}
S.~Chakraborty and J.~Ferdows, ``Lack of hostel facilities and its effects on undergraduate level students’ education: A government college perspective in bangladesh,''
\end{thebibliography}

\end{document}